# Ramsa: A Large Sociolinguistically Rich Emirati Arabic Speech Corpus for ASR and TTS

**Rania Al-Sabbagh**
Department of Foreign Languages
University of Sharjah, UAE
rmalsabbagh@sharjah.ac.ae

**Abstract**

Ramsa is a developing 41-hour speech corpus of Emirati Arabic designed to support sociolinguistic research and low-resource language technologies. It contains recordings from structured interviews with native speakers and episodes from national television shows. The corpus features 157 speakers (59 female, 98 male), spans subdialects such as Urban, Bedouin, and Mountain/Shihhi, and covers topics such as cultural heritage, agriculture and sustainability, daily life, professional trajectories, and architecture. It consists of 91 monologic and 79 dialogic recordings, varying in length and recording conditions. A 10% subset was used to evaluate commercial and open-source models for automatic speech recognition (ASR) and text-to-speech (TTS) in a zero-shot setting to establish initial baselines. Whisper-large-v3-turbo achieved the best ASR performance, with average word and character error rates of 0.268 and 0.144, respectively. MMS-TTS-Ara reported the best mean word and character rates of 0.285 and 0.081, respectively, for TTS. These baselines are competitive but leave substantial room for improvement. The paper highlights the challenges encountered and provides directions for future work.

## 1. Introduction

Gulf Arabic speech corpora remain limited in scale and sociolinguistic detail compared to those available for Indo-European languages and other Arabic varieties (Khalifa et al., 2016; Abid, 2020; Alharbi et al., 2024; Talafha et al., 2024). For Emirati Arabic specifically, recent initiatives, such as Alsanaa (Alblooki et al., 2025), Mixat (Al Ali and Aldarmaki, 2024), ZAEBUC-Spoken (Hamed et al., 2024), Casablanca (Talafha et al., 2024) and ADI17 (Shon et al., 2020), have begun to address the gap. However, limitations in corpus size, speaker representativeness, and the treatment of Emirati Arabic as a homogeneous variety remain.

Regarding corpus size, ADI17 (Shon et al., 2020) is the largest corpus of spoken Emirati Arabic (approximately 112 hours), but it lacks metadata on gender and subdialects. Other corpora are substantially smaller: Alsanaa (Alblooki et al., 2025) contains 4 hours, Casablanca (Talafha et al., 2024) has 6 hours, and Mixat (Al Ali and Aldarmaki, 2024) offers 14.9 hours. Gender imbalance is also prevalent. Alsanaa includes only one male speaker; in Casablanca, only 25.57% of Emirati speakers are women; and ZAEBUC-Spoken includes 26 female speakers and one male speaker. Furthermore, existing corpora treat Emirati Arabic as a homogeneous variety, overlooking well-documented internal variation across Urban, Bedouin, and Mountain/Shihhi subdialects (Obaid, 2006, 2016).

Ramsa is designed to address these gaps. It comprises 41 hours of spoken Emirati Arabic collected from structured interviews and national television shows. It also improves the representation of the female gender with 59 female speakers out of 157 speakers. Furthermore, it reflects subdialectal diversity: the interviewees self-identified as Urban, Bedouin, Mountain/Shihhi or Mixed when a single category was not applicable. The broadcast episodes were selected from shows filmed in locations such as Abu Dhabi, Sharjah, Liwa Oasis, and Al Dhaid, chosen for their associations with different subdialects.

Ramsa recordings also vary in length, interactional format, and recording conditions. Durations are short (<5 minutes), medium (5 to 15 minutes), and long (>15 minutes). Two interactional formats are represented: monologic (a single speaker) and dialogic (exchanges between hosts and guests). Some segments have introductory or background music; most contain little to no background noise.

A 10% subset of Ramsa was used to evaluate commercial and open-source automatic speech recognition (ASR) and text-to-speech (TTS) models in zero-shot settings to establish baseline performance. These benchmarks provide initial reference points and highlight directions for future development.

The contributions of this study are threefold. First, it introduces Ramsa, a 41-hour corpus of spoken Emirati Arabic with broader sociolinguistic representation than existing resources, includ-ing improved female participation and coverage of multiple subdialects. Second, the corpus captures the variation in interactional format, segment length, and recording conditions, providing a resource suitable for diverse speech technology and sociolinguistic applications. Third, it provides initial zero-shot ASR and TTS benchmarks on a subset of the data, establishing baseline performance, and highlighting challenges for future research.

## 2. Related Work

A substantial body of work has focused on dialectal Arabic corpora (e.g., Zaidan and Callison-Burch, 2011; Bouamor et al., 2014; Meftouh et al., 2015; Khalifa et al., 2016; Al-Twairesh et al., 2017; Alsarsour et al., 2018; Zaghouani and Charfi, 2018; Bouamor et al., 2019; Balabel et al., 2020; El-Haj, 2020; Abdelali et al., 2021; Al-Haff et al., 2022; Alkanhal et al., 2023; Jarrar et al., 2023; Nayouf et al., 2023; AlAzzam et al., 2024; El-Ghawi, 2025; Lodagala et al., 2025). This work reflects the increasing demand for dialect-aware speech and text processing resources. Given the scope of Ramsa, this section reviews only speech corpora that include Gulf dialects, with particular attention to Emirati Arabic. For complete reviews of Arabic corpora, the reader may refer to Althobaiti (2020), Awdeh et al. (2021), and Ahmed et al. (2022).

Alblooki et al. (2025) introduced Alsanaa, a 4-hour corpus consisting of a single male speaker reading a Bedouin heritage text. The corpus was used to evaluate several ASR models, including Wav2Vec2.0 (Baevski et al., 2020), XLS-R (Babu et al., 2022), Whisper Small and Medium (Radford et al., 2023), and MMS-TTS-Ara (Pratap et al., 2024). In a zero-shot setting, Wav2Vec2.0 achieved the best performance, with a Word Error Rate (WER) of 46.50 and a Character Error Rate (CER) of 17.13. After fine-tuning, these scores improved to 44.30 WER and 15.96 CER. MMS-TTS-Ara benefited the most from fine-tuning, with WER dropping from 67.21 to 41.04 and CER from 24.56 to 13.34. However, the corpus remains limited to a single speaker and a single subdialect.

Hamed et al. (2024) introduced ZAEBUC-Spoken, an 11.9-hour corpus of role-played Zoom interactions. The dataset includes 27 Emirati students (mainly female), one moderator, and eight interlocutors from Egypt, Europe, and China. Speech features frequent code-switching among Gulf Arabic, Modern Standard Arabic, Egyptian Arabic, and English. Transcriptions were manually produced using the CODA framework (Habash et al., 2018), normalizing dialectal orthography and morphology to Modern Standard Arabic (MSA).

Al Ali and Aldarmaki (2024) compiled Mixat, a 14.9-hour corpus of two Emirati-hosted podcasts that span domains such as sports, finance, science, technology, and health. The episodes were selected for code-switching frequency, and the transcriptions use dialectal spelling rather than normalized orthography.

Talafha et al. (2024) introduced Casablanca, a 48-hour YouTube-based corpus that spans eight dialect regions, including approximately six hours of Emirati speech. Transcriptions were produced manually using dialectal orthography, with Latin scripts used for code-switched items. In zero-shot evaluation on the Emirati subset, Whisper-large-v2 achieved the best results, with a WER of 52.03 and a CER of 19.15, outperforming Seamless-m4t-v2-large (Barrault et al., 2023) and MMS-1B-All (Pratap et al., 2024). After fine-tuning, Whisper-Egyptian performed best on Emirati data (56.58 WER, 20.27 CER). This result may be attributed to the 2,724 lexical items shared between Egyptian and Emirati Arabic, as reported by the authors. However, the corpus remains limited by skewed gender representation, as Emirati women constitute only 25.75% of the speakers, and by the absence of subdialectal variation.

Shon et al. (2020) introduced ADI17, a large-scale dialect identification corpus drawn from more than 3,000 hours of YouTube data in 17 Arab countries, including approximately 112 hours of Emirati Arabic. The corpus is suitable for country-level dialect identification, but lacks sociolinguistic metadata, such as gender and subdialect, which limits its descriptive and analytical value.

These corpora have significantly advanced the documentation and modeling of Gulf Arabic speech. However, they remain limited in the coverage of the subdialect, the richness of metadata, and the female gender representation—gaps that motivate the development of Ramsa.

## 3. Ramsa Sources

Ramsa draws on two sources: (1) structured interviews with native speakers and (2) ten television shows broadcast on national Emirati channels. Details of data availability and distribution are provided in Section 9.

### 3.1. Structured Interviews

The participants were recruited from the University of Sharjah. Ethical approval was secured prior to data collection. Fourteen undergraduate and postgraduate students volunteered to participate: thirteen women and one man. Eligibility required that participants be at least 18 years old, born to Emirati parents, and raised in the UAE.

The interviews were conducted in a quiet office setting, using a noise-reduction microphone to ensure high-quality audio. Each session was recorded as a monologue, with only the participant's voice captured and the interviewer's speech excluded.

| | | |
|---|---|---|
| **Gender** | Male | 1 |
| | Female | 13 |
| **Age** | Millennials | 5 |
| | Gen Z | 9 |
| **Birth Emirates** | Dubai | 3 |
| | Sharjah | 6 |
| | Umm Quwain | 1 |
| | Ras El Khaima | 2 |
| | Fujairah | 1 |
| **Subdialect** | Urban | 3 |
| | Bedouin | 1 |
| | Mountain/Shihhi | 1 |
| | Mixed | 8 |

Table 1: Interviewees' demographics

Before recording, participants completed a short demographic questionnaire covering (1) gender, (2) age range, (3) birth emirate, and (4) self-identified subdialect. For subdialect classification, participants selected Urban, Bedouin, Mountain/Shihhi, or Mixed, allowing for backgrounds that did not align with a single subdialect. Participants were given 30 questions and asked to choose 15 or 20 questions that they felt comfortable answering. The topics included daily routines, food and drink preferences, social customs, and personal interests (see the Appendix A).

As Table 1 shows, there is an imbalance in both gender and subdialect representation, largely due to logistical factors. First, although students and staff were invited by email to volunteer, most of the respondents were from the Department of Foreign Languages of the University of Sharjah, which is predominantly composed of female students. This resulted in a substantial skew toward female participants (13) compared to male speakers (1).

Second, the student population at the University of Sharjah is primarily composed of speakers from Urban or mixed subdialects. The Mountain/Shihhi subdialect is spoken mainly in Ras Al Khaimah, and students from that region do not frequently travel to study at the University of Sharjah—at least not within the Department of Foreign Languages. A similar situation applies to Bedouin students. Furthermore, participants frequently noted that clear-cut boundaries between subdialects are becoming less distinct, particularly among younger speakers whose families may come from different subdialectal backgrounds and who move across the country for study and work. As a result, many speakers report using and being exposed to multiple subdialects through social networks and everyday interaction. This sociolinguistic context is also reflected in the self-reported data, where 8 of the 14 participants identified their native subdialect as a mixed subdialect, most commonly Urban–Bedouin or Urban–Mountain/Shihhi. Addressing these imbalances would require recruiting speakers from multiple disciplines (e.g. Humanities, Science, Technology, Engineering, and Mathematics) as well as from different universities in the United Arab Emirates.

The recordings totaled 2 hours and 43 minutes, with an average interview duration of 11 minutes and 39 seconds.

**3.2. Television Shows**

Several criteria guided the selection of the ten television shows included in Ramsa. All shows were sourced from Emirati broadcasters operated by the national government: Emarat TV, Noor Dubai, Sharjah TV, and Al Wousta Channel. The selected shows typically feature Emirati hosts and guests.

The shows span various topics, including cultural heritage, agriculture and sustainability, professional achievements, cuisine, architecture, and local communities; brief descriptions of each show are provided in Table 2.

The shows have monologic and dialogic formats. The monologic format features a single speaker addressing the audience; the dialogic format involves spontaneous exchanges between hosts and guests. Guests are public figures, including government officials, business leaders, academics, physicians, writers, artists, entrepreneurs, and cultural contributors.

The shows also vary in acoustic quality. Most contain little to no background noise. Some include short musical introductions or closing sequences. Some feature low-level background music during speech.

The shows reflect subdialectal variation. Although most shows represent the Urban subdialect, two shows include speakers associated with the Bedouin subdialect. Hiwayāt wa Muqtanayāt (Hobbies and Collections), broadcast on Al Wousta Channel, highlights the local heritage of Al Dhaid through interviews with local residents. Al Dhaid is predominantly associated with the Bedouin subdialect (Government of Sharjah, 2017). Similarly, Qiṣaṣ min Liwa (Stories from Liwa) features speakers from the Liwa Oasis and emphasizes local heritage; Liwa is also linked to the Bedouin subdialect (Al-Ghāwī, 2012).

At the current stage, the television shows included in Ramsa do not feature speakers from the Mountain/Shihhi subdialect. This variety is primarily associated with Ras Al Khaimah (as mentioned in Section 3.1, and no television shows focusing on this regional variety were identified among the sources considered for the corpus.

| Show Name | Description |
|---|---|
| Sīrat Imraʾa | A biographical interview show featuring Emirati women sharing personal and professional experiences across cultural and occupational domains. |
| Buyūt al-Shāriqa | A heritage documentary exploring Sharjah's historic neighborhoods, architecture, and long-term residents. |
| al-Salfa wī-Mafīhā | A live talk show discussing social issues and community topics through moderated dialogues and audience participation. |
| Hiwayāt wa-Muqtanayāt | A documentary series showcasing individual collections and hobbies that reflect Emirati cultural heritage. |
| Qiṣaṣ min Liwa | A heritage show featuring oral histories and local traditions from Liwa Oasis. |
| Nirwīhā | A documentary highlighting agricultural and livestock initiatives with an emphasis on sustainability and innovation. |
| Lā Khalīna Minkum | A documentary talk show presenting life stories and reflections from elder Emiratis across different regions. |
| Maṭbakh al-Dār | A cooking show featuring conversational interactions around traditional Emirati recipes and domestic practices. |
| Irthunā | An archival documentary preserving oral histories and cultural traditions through first-person narratives. |
| Waṭan al-Riyādah | A documentary interview series chronicling the historical development of key national sectors such as health, media, and education. |

Table 2: Descriptions of television shows included in Ramsa

Table 3 provides descriptive statistics and metadata for each television show included in Ramsa.

## 4. Ramsa Annotations

Four linguistically trained annotators, all graduates of linguistics and translation programs with professional research experience, annotated Ramsa. Three annotators were native speakers of Emirati Arabic—two from Urban backgrounds and one from a Bedouin background—while the fourth annotator was I.

For segmentation, three annotators followed the guidelines outlined in Section 4.1. I served as adjudicator, reviewing the segments, correcting errors, and resolving conflicts. For transcription (Sections 4.2 and 4.3), the Bedouin annotator acted as adjudicator.

### 4.1. Segmentation

Ramsa is segmented into utterances, where an utterance is a single complete communicative act. A complete communicative act is the minimal span of speech that performs a recognizable interactional function (e.g., informing, requesting, or evaluating) and reaches a prosodical or interactional end such that no additional material is needed for it to stand as a usable move (ten Have, 2011).

**Operational criteria**. A span is treated as an utterance if one or more of the following apply:

- **Action fulfillment:** it performs a discourse action (e.g., tell, ask, request, acknowledge, assess, offer, or commit).

- **Closure signal:** it ends with final prosody (e.g., falling contour) or a boundary-relevant pause or reset.

- **Interactional uptake:** it can elicit an immediate response (e.g., an answer, acknowledgment, compliance, or turn shift).

**Examples of utterances in Ramsa include:**

- **Statements** نحنا الحينة في حظيرة المداني (niḥinā il-ḥīnah fī ḥaẓīrit il-midānī, we are now in Al-Midani animal pen)

- **Questions:** الحين هذي شو؟ (al-ḥīn hādhī shū? Now, this is what?)

- **Requests:** زيدي مايي (zīdī māyy, Add more water)

- **Acknowledgments/backchannels:** هيه (hīh, yeah), إنزين (inzīn, alright), صدج؟ (ṣidj? Really?)

- **Elliptical but meaingful utterances:** Q: متى بتردون؟ (matā bitruddūn? When will you be back?) A: عُجب المغرب (ʿugub al-maghrib, after sunset)

**Non-utterances** include:

- **Hesitations or fillers** (e.g., مم... يعني...), which serve only as floor-holding

- **Mid-repair fragments** (e.g., انني—), unless already meaningful in context.

| Show Name | TDiM | NOE | AEL | Time Span | Format | Speakers | | Music | |
| --- | --- | --- | --- | --- | --- | --- | --- | --- | --- |
| | | | | | | M | F | In/Out | BKG |
| Sīrat Imraʾa | 8.18 | 10 | 50 | 07–09/20 | Di. | 0 | 11 | ✗ | ✗ |
| Buyūt al-Shāriqa | 1.09 | 7 | 10 | 07–08/25 | Mono. | 4 | 3 | ✓ | ✗ |
| al-Salfa wī-Mafīhā | 13.41 | 20 | 41 | 06–07/25 | Di. | 11 | 10 | ✓ | ✗ |
| Hiwayāt wa-Muqtanayāt | 3.43 | 15 | 15 | 10/19–01/20 | Di. | 16 | 0 | ✗ | ✗ |
| Qiṣaṣ min Liwa | 0.10 | 4 | 2 | 09/25 | Di. | 6 | 1 | ✗ | ✗ |
| Nirwīhā | 0.44 | 39 | 1 | 04/24–08/25 | Mono. | 25 | 4 | ✓ | ✓ |
| Lā Khalīna Minkum | 2.41 | 9 | 18 | 02–03/23 | Mono. | 9 | 1 | ✓ | ✗ |
| Maṭbakh al-Dār | 5.50 | 18 | 20 | 04/22 | Di. | 0 | 2 | ✓ | ✓ |
| Irthunā | 0.50 | 20 | 2 | 02/23 | Mono. | 17 | 5 | ✓ | ✗ |
| Waṭan al-Riyādah | 1.23 | 17 | 5 | 07–11/22 | Di. | 9 | 9 | ✗ | ✗ |
| **Total** | **38.29** | **159** | | **07/20–09/25** | | **97** | **46** | | |

Table 3: Statistics of television shows in Ramsa. TDiM = total duration in minutes; NOE = number of episodes; AEL = average episode length (minutes); Time Span = period between the earliest and latest episodes included; Format = interaction type (Mono. = monologic, Di. = dialogic); Speakers = counts by gender (M = male, F = female); In/Out = presence of introductory or closing music; BKG = presence of background music.

## 4.2. Dialectal Orthography

Following Al Ali and Aldarmaki (2024), Talafha et al. (2024), Ali et al. (2017), and Wray et al. (2015), Ramsa adopts a dialectal orthographic strategy designed to maintain a close correspondence between transcription and acoustic signal. The goal is to capture the spoken realization as faithfully as possible, not to approximate MSA spelling or to normalize dialectal variation. The following subsections outline the main transcription guidelines adopted in Ramsa.

### 4.2.1. Phonological Reduction

Reductions, assimilations, and clitic fusions are retained without restoring the MSA segmental boundaries. These phenomena are characteristic of spontaneous speech and are transcribed as produced. Examples include

- ما شاء الله (mā shāʾ Allāh, 'God bless') → مشالله (mashāllah)
- هذه الشغلات (hādhihi ash-shaghlāt, 'these things') → هالشغلات (hashshaghlāt)

### 4.2.2. Dialect-Specific Phonological Substitutions

Phonological substitutions are treated as regular phonological features of Emirati Arabic and are transcribed as is. Examples include

- Deletion of the glottal stop: شيء (shayʾ, 'something') → شي (shi)

- /j/ → /y/: جديدة (jadīdah, 'new') → يديدة (yadīdah)
- /q/ → /g/: عقب (ʿuqb, 'after') → عجب (ʿugub)
- /ḍ/ → /ẓ/: بيضا (bayḍah, 'white') → بيظا (bayẓah)
- /k/ → /sh/: حضرتك (ḥaḍratik, 'you.f.sg') → حظرتش (ḥaẓratish)

### 4.2.3. Variation as Produced

When a lexical item has multiple spoken realizations—either within or across speakers—each instance is transcribed according to its actual production, without imposed standardization. For example, the first-person plural pronoun is sometimes pronounced نحن (niḥn) and some other times نحنا (niḥna). Similarly, the temporal adverb now is sometimes pronounced as الحين (ilḥīn) and some other times as الحينة (ilḥīna). Even discourse markers can have different pronunciations as speakers may say زين (zīn, 'well') or إنزين (inzīn).

## 4.3. Other Transcription Guidelines

Other transcription guidelines, adapted from Hamed et al. (2020, 2024), are as follows:

**Punctuation and Numbers.** Annotators apply punctuation at their discretion. Numbers are written in letters rather than digits.

**Unclear or Partially Audible Speech.** If a word is unclear, the annotator replays the audio and provides a best-guess transcription enclosed in double parentheses ((word)). If no plausible guess can be made, only the parentheses remain.

| Source | Durations | | Gender | | Words | | | | Utterances | |
|---|---|---|---|---|---|---|---|---|---|---|
| | TDiM | 10% | M | F | Tokens | Types | TTR | CS | Total | AUL |
| Interviews | 161 | 16 | 1 | 0 | 1,890 | 912 | 0.48 | 1 | 227 | 8 |
| Sīrat Imraʾa | 498 | 50 | 0 | 2 | 5,507 | 2,444 | 0.44 | 13 | 359 | 15 |
| Buyūt al-Shāriqa | 69 | 7 | 1 | 0 | 606 | 379 | 0.63 | 0 | 50 | 12 |
| al-Salfa wī-Mafīhā | 821 | 82 | 2 | 1 | 11,349 | 4,959 | 0.44 | 44 | 1,030 | 11 |
| Hiwayāt wa-Muqtanayāt | 223 | 22 | 2 | 0 | 2,298 | 1,155 | 0.50 | 0 | 202 | 11 |
| Qiṣaṣ min Liwa | 10 | 1 | 2 | 0 | 150 | 125 | 0.83 | 0 | 11 | 14 |
| Nirwīhā | 44 | 4 | 1 | 1 | 282 | 203 | 0.72 | 0 | 27 | 10 |
| Lā Khalīna Minkum | 161 | 16 | 1 | 0 | 1,895 | 1,156 | 0.61 | 0 | 273 | 7 |
| Maṭbakh al-Dār | 350 | 35 | 0 | 2 | 3,326 | 1,859 | 0.56 | 0 | 724 | 5 |
| Irthunā | 50 | 5 | 2 | 1 | 601 | 432 | 0.72 | 0 | 74 | 8 |
| Waṭan al-Riyādah | 83 | 8 | 1 | 2 | 1,397 | 816 | 0.58 | 0 | 119 | 12 |
| **Total** | **2,470** | **246** | **13** | **9** | **29,301** | **11,207** | **0.38** | **57** | **3,096** | **9.5** |

Table 4: Descriptive statistics for the transcribed 10% Ramsa subset. TDiM = total duration in minutes and 10% sampled duration; Gender distribution (M = male, F = female); token and type counts; TTR = type–token ratio; CS = number of code-switched words (written in Latin script); AUL = average utterance length (in words).

**Repetitions, Repairs, and Errors.** All disruptions of fluent speech (e.g., repetitions, false starts, self-repairs, hesitations, or truncated words) are transcribed as produced. Unfinished words are marked with a double hyphen (–); pauses or mid-utterance hesitations are indicated by double dots (..); the same marker is used at the end of a segment when an utterance is suspended or abandoned.

**Non-Speech Sounds.** Non-verbal vocalizations or environmental sounds are enclosed in curly braces, e.g., {laugh}.

**Interjections.** Interjections are prefixed with a percentage sign (٪.), e.g., ٪.هاا.

**Interruptions.** Points where one speaker cuts off another are marked with a tilde (~).

**Script Selection for Arabic vs. English Words.** Arabic words are transcribed in Arabic script. English words are transcribed in Latin script. However, loanwords adapted to Arabic phonology (e.g., أوكيه (okay)) are transcribed in Arabic script.

## 5. Ramsa Statistics

Ramsa is currently a work in progress; at the time of writing this paper, only 10% of the corpus has been manually transcribed. This initial subset was transcribed to support the development of transcription guidelines, identify transcription challenges, train transcribers, and enable preliminary benchmarking.

The 10% sample was obtained by randomly selecting approximately 10% of the material from each source. This sampling strategy allows for preliminary linguistic analysis without privileging any particular genre, topic, or interactional format. The transcribers involved in this phase are described in Section 4, and descriptive statistics for the transcribed subset are provided in Table 4.

In total, 246 minutes of speech (approximately four hours) were transcribed, comprising recordings from 13 male and 9 female speakers. The sample contains 29,301 tokens, 11,207 word types, and 3,096 utterances, with an average utterance length of 9.5 words.

Television shows with higher type–token ratios (TTR), particularly documentary and informational formats, exhibit topic-dense discourse and correspondingly greater lexical diversity. In contrast, conversational, cooking, and talk-show formats display lower TTR values due to frequent use of interactional markers, scaffolding expressions, and repeated referential content.

Some shows also contain many utterances relative to total tokens, reflecting dense turn-taking, short turns, and rapid interactional sequencing—patterns most evident in talk-show and cooking formats. This quantitative profile is expected to evolve as transcription continues, and additional episodes are incorporated. Table 5 presents illustrative transcript excerpts.

## 6. Ramsa Benchmarking

As noted in Section 1, Ramsa is benchmarked on two core speech tasks: automatic speech recognition (ASR) and text-to-speech (TTS), using commercial and open-source models. All results reported in the following subsections are based on the 10% transcribed subset described in Section 5.

### 6.1. ASR

#### 6.1.1. Models

Three models are benchmarked: the commercial systems AssemblyAI Universal-2 and Gladia

| Show Name | Transcription Sample |
|---|---|
| Sīrat Imraʾa | الله يسلمش. // سعيدين جدا انش ويانا اليوم. // ونتناول سيرة شابة إماراتية // نحتت كما ذكرنا سيرة رائعة وجميلة // لا بد أنها ستذكر في تاريخ دولة الإمارات بإذن الله. // لكن دائماً أنا أقول، سعاتش، الواحد عشان يعرف شو سبب هالثمار اليانعة اللي الكل يشوفها دائماً هناك جذور // ونحب دائماً نحن في سيرة امرأة نرجع للجذور. // يا ريتش سعاتش تخبرينا عن البداية، عن الطفولة، البيئة العائلية. // كيف هذه الشخصية نمت في هذه البيئة؟ // اتفظلي // |
| Buyūt al-Shāriqa | بيتنا اللي كان بالشرق، ما شاء الله، بيت كبير. // فالوالد من حبه لخواته إنه نحمس، حتى خواته اعطاهم أجزاء من ال-- من البيت // فبيتنا، ما شاالله، نحنا تخصيا يتكون من جج-- أربع غرف يعني ومطبخ ومجلس كذلك وأدب، // في أدبين البيت ما شاالله، // وخريطة موجودة كذلك في البيت في الوسط ./اه ./اه ./اه // وكذلك بيت عماتي، // أنا وجدتي كلنا احنا متلاصقين مع بعظ // ما ينا ./آه مفتوح البيت، يعني، ./اه، ./اه ما شاء الله. نحنا فرجينا كبير مش صغير في الشرق // ونحنا كل الفريج يعتبر جيران. // |
| Nirwīhā | أول مايه، أشوف عطايا الله. // أشوف، // انطالع، // انطالع عليهم. // انطالع، اللي تعشت، // اللي ما تعشت، // شو فيهن، ما فيهن. // وإذا عندي تعليمات، خبرت عمالي. // وأمر على شبت الغنم. // الحلال، اللي هو من ذاك. // انطالع، // شو زاد عندي؟ // شو اللي تعشي؟ // اللي ما تعشى؟ // شو مريض؟ // شو ولد عندي؟ // ما ولد؟ // إذا كان عندي مواليد جدد. // إهنيه، نحنا الحينة، في حظيرة المداني والمظاري // وعندي من الحلال الأصايل، بنات ظبيان، بنات هملول، بنات الخبارة // وعندي أأصل البوش. // ويوم يشوف حلالي، يفز جلبي. // الواحد يوم يشوف عطايا الله، يسده. // ويشوف السيح. // |

Table 5: Transcription excerpts from three sources, with // marking utterance boundaries.

AI's Solaria/Whisper-Zero, and the open-source Whisper-large-v3-turbo.

AssemblyAI Universal-2 is a proprietary Conformer–RNN-T architecture trained on more than 12 million hours of multilingual audio (Loeber, 2024). It includes an integrated neural text-formatting module (Universal-2-TF) for punctuation, casing, and inverse text normalization (Khare et al., 2025). The model is marketed as optimized for conversational and noisy speech, with enhanced handling of alphanumerics and rare words.

Gladia AI offers two models: Solaria and Whisper-Zero. Solaria provides low-latency multilingual transcription—reportedly below 300 milliseconds interruption latency—and supports over 100 languages, including code-switched input (Gladia AI, 2025). Whisper-Zero is an enterprise adaptation of OpenAI's Whisper, further trained on approximately 1.5 million hours of additional data. Gladia reports relative WER improvements of 10–15% over Whisper-large-v2/v3, with added functionality for streaming, diarization, and translation (Gladia AI, 2024).

Whisper-large-v3-turbo (Radford et al., 2022) is an open-source, 1.55-billion-parameter Transformer encoder–decoder trained on more than 5 million hours of labeled data. The turbo variant is a fine-tuned, pruned version of Whisper-large-v3 with reduced decoding layers (from 32 to 4), offering substantially faster inference while maintaining strong zero-shot generalization across domains.

### 6.1.2. Settings and Evaluation Metrics

All audio files were in MP3 format with standard sampling rates, as obtained from broadcast and interview sources. To ensure consistency, files were submitted to each API in their original format without re-encoding. AssemblyAI Universal-2 used automatic language detection with default parameters, while Gladia AI's Solaria/Whisper-Zero was set to Arabic ("ar") under default settings. No additional preprocessing, segmentation, or speaker diarization was applied. Whisper-large-v3-turbo was evaluated with default inference settings—beam search decoding (beam size = 5), automatic language detection, and 30-second timestamped segments—and handled punctuation and casing natively during decoding. For evaluation, outputs were normalized by removing punctuation to ensure comparability across systems, and performance was measured using word error rate (WER) and character error rate (CER) at the document level.

### 6.1.3. Results

Table 6 reports ASR results on the 10% transcribed subset. Whisper-large-v3 achieves the best average scores (WER 0.268, CER 0.144), outperforming AssemblyAI (WER 0.354, CER 0.175) and Gladia (WER 0.347, CER 0.175). Performance varies by source: interview and documentary-style speech is easier (e.g., Sīrat Imraʾa, Qiṣaṣ min Liwa), whereas overlap-rich conversational formats are hardest—Matbakh al-Dār4 shows dense, rapid turn-taking with frequent over-laps and interruptions, corresponding to the highest error rates across systems (WER ≈ 0.73–0.79).

Compared with prior dialectal benchmarks, Ramsa zero-shot results are stronger for Emirati Arabic than Casablanca (Whisper-large-v2: WER 0.52, CER 0.19; (Talafha et al., 2024)) and comparable to Alsanaa (Wav2Vec 2.0 zero-shot: WER 0.47, CER 0.17; (Alblooki et al., 2025)). As a point of comparison beyond zero-shot settings, the Munsit at NADI 2025 system achieves WER ≈ 0.28 and CER ≈ 0.12 on Saudi Arabic after weakly supervised fine-tuning (Salhab et al., 2025), illustrating the improvement potential with targeted adaptation.

As expected, CER remains lower than WER across systems; character-level metrics are especially informative under dialectal orthography, which reflects surface phonetics and reduces tokenization effects—a trend noted for non-standardized languages with dialectal variation like Swiss German (Nigmatulina et al., 2020) and morphologically rich languages like Arabic (K et al., 2025). Overall, Ramsa establishes a robust Emirati-dialect baseline and a foundation for fine-tuning. For broader benchmarking trends, see Talafha et al. (2024), Wang et al. (2025), Dhouib et al. (2022), and Besdouri et al. (2024).

## 6.2. TTS

### 6.2.1. Models

Two open-source TTS systems were benchmarked: ArTST and MMS-TTS-Ara. ArTST (Arabic Text and Speech Transformer; (Toyin et al., 2023)) follows the SpeechT5 unified-modal framework: a Transformer encoder–decoder that supports both text and speech tasks within a shared architecture (pretraining across modalities, task-specific fine-tuning for TTS). The initial release targets MSA and is designed to be extended to dialectal and code-switched Arabic.

MMS-TTS-Ara (Pratap et al., 2024) is the Arabic checkpoint from Meta's Massively Multilingual Speech (MMS) project. MMS-TTS uses VITS—an end-to-end, adversarially trained, conditional variational autoencoder with a posterior encoder, conditional prior, and neural vocoder—producing waveforms directly from text without an external vocoder. MMS provides per-language TTS models (1,000+ languages), including Arabic, trained with large-scale weakly supervised data.

### 6.2.2. Settings and Evaluation Metrics

All models were evaluated using default configurations. Both ArTST and MMS-TTS-Ara were run without parameter modification or speaker-embedding customization; in ArTST, the default automatic speaker embedding of the model was retained. The input text comprised the gold-standard Ramsa transcriptions rather than ASR-generated text, ensuring that synthesis quality was assessed independently of recognition performance. To evaluate intelligibility and pronunciation accuracy, the generated audio was back-transcribed using Whisper-large-v3-turbo in zero-shot mode, and the resulting transcriptions were compared with the reference text after punctuation normalization for consistency. Evaluation used WER and CER at the document level, following the standard back-transcription approach adopted in TTS benchmarking (Liu et al., 2025; Pratap et al., 2024).

### 6.2.3. Results

Table 7 reports WER/CER by source. On average, MMS-TTS-Ara outperforms ArTST (WER 0.285, CER 0.081 vs. WER 0.373, CER 0.161). The same trend holds across individual shows: MMS yields lower error rates for interviews and informational or documentary speech (e.g., Qiṣaṣ min Liwa: WER 0.153, CER 0.049), while overlap-rich, rapid-turn formats remain challenging for both systems (e.g., Maṭbakh al-Dār: ArTST WER 0.579, CER 0.281; MMS WER 0.455, CER 0.127). As in the ASR results, CER values remain consistently lower than WER, reflecting reduced sensitivity to tokenization and closer alignment with surface phonetics under dialectal orthography.

Zero-shot and multidialectal Arabic TTS remain underexplored. In Doan et al. (2024), XTTS models fine-tuned on ≈472 hours of Arabic report performance on the QASR benchmark (Mubarak et al., 2021), using metrics such as ASR-WER and Speaker Embedding Cosine Similarity (SECS). On unseen QASR speakers, the baseline model attains WER ≈ 6.4 and SECS ≈ 0.755, while dialect-specific fine-tuning increases WER (≈ 16–18) but improves SECS (≈ 0.766). Importantly, QASR does not include Emirati Arabic, and—to the best of my knowledge—no published TTS results currently exist for Emirati speech.

For comparison, Egyptian Arabic TTS systems such as Masry achieve high Mean Opinion Scores (MOS) under controlled, single-speaker studio conditions. For example, Tacotron-based models using Griffin–Lim and HiFi-GAN vocoders report MOS values of approximately 4.48 for Tacotron 2 and 3.64 for Tacotron 1 (Azab et al., 2023). However, such controlled experimental setups differ markedly from Ramsa's multi-speaker, broadcast-style data and from my zero-shot, ASR-based evaluation protocol.

Overall, Ramsa TTS baselines provide the first replicable, zero-shot reference for Emirati Arabic. MMS-TTS-Ara consistently surpasses ArTST in objective intelligibility, with the most significant low

| Source | AssemblyAI | | Gladia | | Whisper-Large-v3 | |
|---|---|---|---|---|---|---|
| | WER | CER | WER | CER | WER | CER |
| Interviews | 0.362 | 0.168 | 0.403 | 0.203 | 0.272 | 0.151 |
| Sīrat Imraʾa | 0.211 | 0.082 | 0.210 | 0.087 | 0.138 | 0.054 |
| Buyūt al-Shāriqa | 0.236 | 0.088 | 0.258 | 0.106 | 0.217 | 0.108 |
| al-Salfa wī-Mafīhā | 0.322 | 0.166 | 0.211 | 0.095 | 0.203 | 0.112 |
| Hiwayāt wa-Muqtanayāt | 0.443 | 0.305 | 0.248 | 0.095 | 0.181 | 0.084 |
| Qiṣaṣ min Liwa | 0.213 | 0.092 | 0.207 | 0.062 | 0.173 | 0.080 |
| Nirwīhā | 0.213 | 0.073 | 0.488 | 0.361 | 0.141 | 0.049 |
| Lā Khalīna Minkum | 0.397 | 0.160 | 0.407 | 0.170 | 0.345 | 0.161 |
| Maṭbakh al-Dār | 0.790 | 0.499 | 0.773 | 0.483 | 0.727 | 0.534 |
| Irthunā | 0.443 | 0.181 | 0.444 | 0.190 | 0.373 | 0.176 |
| Waṭan al-Riyādah | 0.266 | 0.108 | 0.164 | 0.068 | 0.178 | 0.077 |
| **Averages** | **0.354** | **0.175** | **0.347** | **0.175** | **0.268** | **0.144** |

Table 6: ASR baseline results

| Source | ArTST | | MMS | |
|---|---|---|---|---|
| | WER | CER | WER | CER |
| Interviews | 0.336 | 0.133 | 0.328 | 0.104 |
| Sīrat Imraʾa | 0.349 | 0.164 | 0.238 | 0.066 |
| Buyūt al-Shāriqa | 0.359 | 0.186 | 0.266 | 0.076 |
| al-Salfa wī-Mafīhā | 0.335 | 0.159 | 0.245 | 0.074 |
| Hiwayāt wa-… | 0.297 | 0.134 | 0.229 | 0.058 |
| Qiṣaṣ min Liwa | 0.273 | 0.099 | 0.153 | 0.049 |
| Nirwīhā | 0.368 | 0.106 | 0.253 | 0.055 |
| Lā Khalīna Min… | 0.422 | 0.179 | 0.329 | 0.099 |
| Maṭbakh al-Dār | 0.579 | 0.281 | 0.455 | 0.127 |
| Irthunā | 0.453 | 0.176 | 0.413 | 0.114 |
| Waṭan al-Riyādah | 0.336 | 0.151 | 0.228 | 0.068 |
| **Averages** | **0.373** | **0.161** | **0.285** | **0.081** |

Table 7: TTS baseline results

results observed in overlap-dense sources.

## 7. Conclusion and Outlook

Ramsa is a developing large-scale speech corpus of Emirati Arabic designed to support both computational and sociolinguistic research. Transcription is ongoing, and the ASR and TTS baseline results presented in this study are based on a representative 10% subset of the corpus. Yet, the results reveal consistent cross-model patterns: recognition accuracy is highest for monologic and documentary speech and lowest for overlap-rich conversational formats.

While the design of Ramsa substantially improves gender balance and subdialectal coverage for Emirati Arabic, these same dimensions also introduce challenges that future work will address. Access to Urban Emirati speech is comparatively strong, whereas both source material and qualified annotators remain scarce for the Bedouin and Mountain/Shihhi subdialects. This imbalance affects the scope and interpretation of the current ASR and TTS benchmarks and constrains evaluation options. For example, a Mean Opinion Score (MOS) assessment, a standard evaluation method for speech synthesis, could not be conducted because expert listeners representing all subdialects were not available for this study.

In addition, a generational dimension emerged during annotation. Recruited annotators fell within the 20–40 age range, and some reported difficulty interpreting some lexical items. This suggests ongoing language change and possible attrition of certain forms, indicating that future annotation efforts would benefit from including speakers and annotators from a wider range of age groups. Addressing these challenges through expanded data collection, diversified annotation expertise, and broader evaluation protocols will further strengthen Ramsa as a resource for Emirati Arabic sociolinguistic, ASR, and TTS research.

## 8. Limitations

The work presented in this paper represents the initial stage of the Ramsa project, as corpus development and transcription remain ongoing. Consequently, the current results—particularly for ASR and TTS—should be interpreted as preliminary benchmarks, subject to revision as the dataset grows and annotation consistency improves.

At this stage, approximately 10% of the total audio has been manually transcribed, constraining the scope of both linguistic and technological analyses. This limited coverage may affect the representativeness of the baseline results and reduce the robustness of model evaluations across speakers, topics, and recording conditions.

Although Ramsa improves female speaker representation relative to prior Emirati corpora, overall gender balance remains uneven. Also, the corpus primarily reflects Urban Emirati speech, with limited representation of Bedouin and even less of Mountain/Shihhi speakers. These demographics and subdialectal asymmetries may influence linguistic generalizations and system performance, particularly for dialect-sensitive models.

Finally, all reported benchmarks reflect zero-shot performance without model fine-tuning or domain adaptation. The current results thus represent the generalization capacity of existing multilingual models, rather than the upper bound of performance achievable through targeted adaptation or fine-tuning on Emirati Arabic.

# 9. Ethical Statement: Data Availability and Distribution

## 9.1. Interview Data (Restricted Access)

For the interview component, participants provided their informed consent that allowed research use under restricted access conditions. Audio recordings and associated transcripts (with pseudonymized speaker identifiers) are available on request to qualified researchers for non-commercial scholarly use, subject to an institutional Data Use Agreement (DUA) approved by the university's ethics board.

**The DUA:**

- Prohibits redistribution, secondary sharing, or attempts at re-identification
- Requires secure storage, access controls, and restricted personnel access
- Limits use to the approved research purpose
- Requires the designation of responsible personnel

**Qualified researchers must:**

- Be affiliated with a recognized research institution
- Provide documentation of ethics/IRB approval or formal exemption covering the proposed use
- Agree to non-commercial use only
- Demonstrate appropriate data security safeguards
- Execute the institutional DUA

Where access involves transfer outside the UAE, applicants must confirm implementation of a legally compliant cross-border transfer mechanism and adherence to applicable data protection obligations, including respect for participants' data-subject rights.

## 9.2. Broadcast Component (Copyright-Restricted)

For the broadcast component, television audio or video recordings and full verbatim transcripts are not distributed due to copyright and platform-licensing constraints.

To support transparency and replicability, the following materials are provided upon request:

- Official source URLs as made publicly available by the rightsholder
- Non-expressive metadata (show title, channel, broadcast date, episode or segment identifiers, and host/guest names as credited on air)
- Analytical coding files, evaluation scripts, and methodological documentation

We do not circumvent technical protection measures or access controls, nor do we redistribute platform-provided subtitles or captions (where applicable). All copyrights and related rights remain with the respective rightsholders.

## A. Appendix A: Interview Prompts

الحياة اليومية والعادات

- صف لي كيف تبدأ يومك من اللحظة التي تستيقظ فيها حتى خروجك من المنزل.
- ما هي العادات الصغيرة التي لا تستغني عنها في يومك العادي؟
- ما هي العادات الصغيرة التي لا تستغني عنها في يومك العادي؟
- كيف يختلف يوم عطلة نهاية الأسبوع عن باقي أيامك؟
- احكِ لي عن موقف تتذكره غيّر روتينك اليومي.
- ما هي الأماكن التي تزورها بشكل متكرر في حياتك اليومية، ولماذا؟

الطعام والشراب

- صف لي وجبتك المفضلة: كيف تُحضّر، ومتى تحب تناولها أكثر؟
- ما هي ذكرياتك مع وجبة مرتبطة بالطفولة أو العائلة؟
- احكِ لي عن مطعم أو مقهى يعجبك: ماذا يميّزه بالنسبة لك؟
- ما هو طبق تقليدي تعتقد أنه يمثل هويتكم؟ كيف يوصف للذي لم يجرّبه من قبل؟
- إذا كنت تحب تجربة أكلات جديدة، ما آخر شيء جربته وكيف كان شعورك؟

الطقس والأجواء

- صف لي يوماً شديد الحرارة: كيف تقضي وقتك، وما الأنشطة التي تناسب هذا الجو؟
- كيف يغيّر فصل الشتاء من عاداتك اليومية؟
- ما هو المكان المفضل لديك للهروب من حرارة الصيف، وما الذي يجعلك ترتاح فيه؟
- احكِ لي عن تجربة جميلة تتذكرها مع نزول المطر.
- كيف يختلف شعورك وأنشطتك في البحر عن الصحراء أو الجبال؟

الهوايات ووقت الفراغ

- ما هي الهواية أو النشاط الذي يشغلك أكثر وقت فراغك؟ صف كيف تمارسه.
- إذا قضيت وقتاً ممتعاً مع أصدقائك، كيف يكون شكل هذا اللقاء عادة؟
- صف لي فيلماً أو برنامجاً تلفزيونياً أثّر فيك، وما الذي أعجبك فيه؟
- كيف ترى دور وسائل التواصل الاجتماعي في حياتك اليومية؟
- ما هي الأنشطة التي تفضّل أن تقوم بها وحدك، وما هي التي تفضّل أن تشاركها مع الآخرين؟

العمل والدراسة والمهام اليومية

- كيف يبدو يومك المعتاد في العمل أو الدراسة من بدايته إلى نهايته؟
- ما هو الموقف الأكثر إثارة أو صعوبة واجهته في عملك أو دراستك، وكيف تعاملت معه؟
- صف لي طريقة تنظيمك ليوم مهم، مثل امتحان أو اجتماع.
- ما هي الأشياء التي تساعدك على التركيز والإنتاجية؟
- كيف تغيّر يومك بعد انتهاء العمل أو الدراسة، وما العادات التي تساعدك على الاسترخاء؟

المجتمع والحياة الاجتماعية

- صف لي اجتماعاً عائلياً مميزًا حضرته: ماذا فعلتم، وكيف كان الجو؟

- كيف تتغير مواضيع الحديث عندما تجلس مع الأقارب مقارنة بالحديث مع الأصدقاء؟
- ما هو المكان الذي تفضّل لقاء أصدقائك فيه، وكيف يكون شكل الجلسة؟
- احكِ لي عن عادة اجتماعية تراها جزءًا أساسيًا من حياتكم اليومية.
- صف لي موقفًا مضحكًا أو لطيفًا حدث معك في مناسبة اجتماعية.